\documentclass[conference]{IEEEtran}
\IEEEoverridecommandlockouts
\usepackage{cite}
\usepackage{amsmath,amssymb,amsfonts}
\usepackage{algorithmic}
\usepackage{graphicx}
\usepackage{textcomp}
\usepackage{xcolor}
\usepackage{subfig}
\def\BibTeX{{\rm B\kern-.05em{\sc i\kern-.025em b}\kern-.08em
    T\kern-.1667em\lower.7ex\hbox{E}\kern-.125emX}}
\begin{document}

\title{Defect Detection from UAV Images based on Region-Based CNNs
}

\author{
\IEEEauthorblockN{ Meng Lan, Yipeng Zhang, Lefei Zhang, Bo Du}
\IEEEauthorblockA{School of Computer Science, Wuhan University, Wuhan, China} 
{\{menglan, yp91, zhanglefei, remoteking\}@whu.edu.cn}
}
\maketitle
\footnote {This work was supported by the National Natural Science Foundation of China under Grants 41431175, 61471274, 61771349, and 61711530239.}
\begin{abstract}
With the wide applications of Unmanned Aerial Vehicle (UAV) in engineering such as the inspection of the electrical equipment from distance, the demands of efficient object detection algorithms for abundant images acquired by UAV have also been significantly increased in recent years. In computer vision and data mining communities, traditional object detection methods usually train a class-specific learner (e.g., the SVM) based on the low level features to detect the single class of images by sliding a local window. Thus, they may not suit for the UAV images with complex background and multiple kinds of interest objects. Recently, the deep convolutional neural networks (CNNs) have already shown great advances in the object detection and segmentation fields and outperformed many traditional methods which usually been employed in the past decades. In this work, we study the performance of the region-based CNN for the electrical equipment defect detection by using the UAV images. In order to train the detection model, we collect a UAV images dataset composes of four classes of electrical equipment defects with thousands of annotated labels. Then, based on the region-based faster R-CNN model, we present a multi-class defects detection model for electrical equipment which is more efficient and accurate than traditional single class detection methods. Technically, we have replaced the RoI pooling layer with a similar operation in Tensorflow and promoted the mini-batch to 128 per image in the training procedure. These improvements have slightly increased the speed of detection without any accuracy loss. Therefore, the modified region-based CNN could simultaneously detect multi-class of defects of the electrical devices in nearly real time. Experimental results on the real word electrical equipment images demonstrate that the proposed method achieves better performance than the traditional object detection algorithms in defect detection.
\end{abstract}


\section{Introduction}
With the developments of the sensors and Unmanned Aerial Vehicle (UAV) technologies, the UAV data analysis algorithms have attracted growing attentions in computer vision and data mining communities. Among which, the UAV electrical equipment inspection is an important and challenging task for most of power grid companies. On the one hand, the UAV inspection has the advantages of low risk, low cost and flexible operation compares to the traditional human labor working, however, if the huge amount of data taken by the UAV need to be manually reviewed to get the final inspection report, the results of which may subjective and inefficient in practice. Therefore, it is significant and urgent to develop some advanced algorithms to detect the defects in these UAV images with high accuracy and efficiency.

In the past years, there are a few studies focus on the defect detection using the UAV images for the electrical equipment, among which the major researches are focused on insulators and power transmission line\cite{1}, \cite{2},\cite{3}.According to the exists literature, the related methods may be classified into the following two categories, the traditional object detection and the edge detection.The traditional object detection based methods \cite{4},\cite{5}, \cite{6}, \cite{7} firstly train a classifier for each of the specific categories, and then the trained classifier slides a fixed size window on the same feature representation of the original image to obtain a response score map, in which the location with the larger score may has the higher probability to become the interest object. In particular, \cite{8} proposes the SIFT feature for the identification and location of the insulators. On the other hand, the edge detection based methods \cite{9}, \cite{10}use the edge detection operators to get the edge information of objects and further utilize this information to achieve a specific task. For example, \cite{11} uses the canny operator to obtain the edge information of the water droplets on the surface of the insulator, and then the BP neural network is used to predict the water repellency of the insulator. All the above mentioned methods have good performances for certain class of electrical equipment, however, most of the exist methods haven’t well considered the interfere of the complex environments in real-world, therefore, these methods may not meet the requirements of detecting the defects of various electrical equipments. 

Recently, the deep learning algorithms have becoming popular in many computer vision and data mining related applications, in particular, the deep convolutional neural networks (CNNs) have achieved the state-of-the-art performance in image classification, segmentation, recognition, and detection tasks \cite{12},\cite{13}, \cite{14}, \cite{15}. Among which, the deep learning based object detection methods mainly study the common types of natural images such as people, car and some commonly see animals, and the interested objects in the training datasets are always both rich in number and significant in size. However, in our UAV based electrical equipment defect detection task, the defects to be detected in the images are very small and even inconspicuous, which increases the difficulty of defects detection. Considering the fact that the UAV images are also visual image and the region-based object detection algorithms have good properties for detecting small objects by using the mechanism of anchors and feature pyramid network (FPN)\cite{16} and precise label, therefore, it may possible to apply the deep CNNs to the defect detection of electrical equipment, i.e., the problem we have now. In fact, there are already exist a few researches in deep learning based defect detection of electrical equipment. For instance, \cite{17} uses deep CNN for Status Classification of power line insulator and \cite{18} uses deep CNN for detection of transmission line. However, these methods mainly employ the deep CNN to extract the feature and are only designed for specific electrical equipment.

In this work, we introduce the region-based object detection algorithm, i.e., faster R-CNN to detect the defects of multi-class electrical equipment using the UAV images. Note the fact that the input UAV image has the properties of high spatial resolution, complex backgrounds and low contrast between objects and background, we preprocess the images before they are put into the network. In addition, we slightly modify the faster R-CNN to fit our problem in practice. First, following the finding in \cite{19}, instead of using the RoI pooling layer, we use the crop\_and\_resize operation, which crops and resizes feature maps to 14 x 14, and then max-pool them to 7 x 7 to match the input size of fc6. Second, the original faster R-CNN \cite{20} removes small proposal which is less than 16 pixels in height or width in the original scale resulting in negative performance for small objects and we preserve all the small region proposals in our modified network. The final experimental results show that the proposed method can simultaneously detect various defects of electrical equipment and achieves better performance than the traditional object detection algorithms in defect detection. In summary, the main contributions of our work can be summarized as follows:
\begin{itemize}
\item A region based multi-class defective electrical equipments detection method is introduced for the efficient and accurate inspection system.
\item A multiple classes of electrical equipment dataset with precise label has been built from more than five thousands of UAV HD images.
\item The structure of the region based CNN has been improved for faster detection while keeping the accuracy.
\end{itemize}

The remainder of this paper is organized as follows. Section 2 introduces two research direction of object detection. Section 3 proposes the modified region-based R-CNN for multiple class defects detection .In section 4, we evaluate the model and the results demonstrate that our model has a good performance. Finally, conclusion and feature work are presented in section 5.

\section{RELATED WORK}

Before the deep learning era, the tranditional object detection methods \cite{4},\cite{5}, \cite{6}, \cite{7} follow the basic pipeline of pattern recognition, which usually composed of two parts, i.e., feature extraction and classifier training. In the last decade, the most widely used features are the SIFT and HOG which are blockwise orientation histograms. The SVM and its derivatives are frequently trained as the classifier to predict the possible location of the object by sliding a window over the whole image. These traditional methods may get good performance for some simple tasks, but their accuracy is not satisfied to meet the need of advanced visual tasks especially the multi-class object detection in practice.

Around the year 2014, CNN is firstly employed for feature extraction in the region-based object detection algorithm. The R-CNN \cite{21} trains the deep CNNs to extract a fixed-length feature vector from region proposals and then that feature vector is input into a set of class-specific linear SVMs to classify object categories or background. However, it is known that the R-CNN could not predict object bounds and its accuracy depends on the performance of the separate region proposals module such as selective search \cite{22}. To relieve this issue, using deep neural networks to predict object bounding box have been studied in several other works \cite{23}, \cite{24}, \cite{25}, \cite{26}. To name a few, the Overfeat method \cite{23} trains a fully-connected layer to predict the bounding box coordinates for single object, and then the fc layer is turned into a convolutional layer for detecting multiple objects.The MultiBox methods \cite{25}, \cite{26} generate region proposals from a network whose last fully-connected layer simultaneously predicts multiple class-agnostic boxes, and these class-agnostic boxes are used as proposals for R-CNN.

\section{MODIFIED REGION-BASED CNN FOR UAV IMAGES DEFECT DETECTION}
\begin{figure}
\centering
\includegraphics[height=4cm,width=0.45\textwidth]{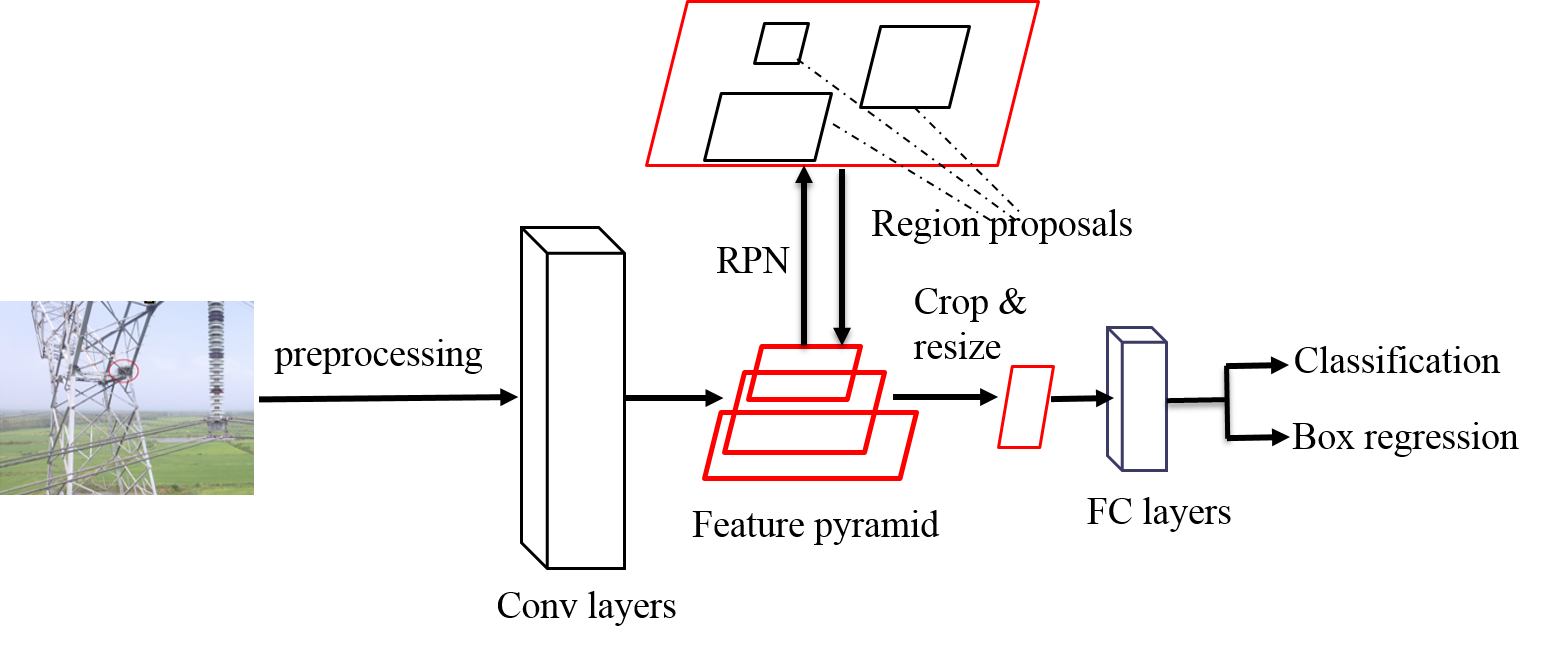}
\caption{\small Architecture of he proposed method. The model composed of region proposals network and fast R-CNN which sharing the deep convolutional layers.}
\label{fig1}
\end{figure}

The object detection algorithm faster R-CNN is composed of two modules.The first module is a deep fully convolutional network \cite{28} that proposes region proposals, and the second module is the fast R-CNN detector \cite{14} which takes as input the region proposals and images. Through sharing the features of deep convolution layers, the whole algorithm becomes a single and unified network for object detection, shown in Figure \ref{fig1}. 

\subsection{UAV Data Preprocessing}\label{AA}
To well train the defect detection model, we have collected a lot of HD UAV images without labels, and then manually labeled thousands of images. However, it is still not sufficient enough to train a precise detector model for multi-class electrical equipment in practice. In the literature, the most common method to reduce overfitting in CNNs training step is to artificially enlarge the dataset using label-preserving transformations \cite{12}, \cite{29}.

In this study, the first method of data augmentation we considered is horizontal reflection. This trick doubles the size of our training set and the transformed images could be produced from the original images with very few computational costs, in addition, these transformed images do not need to be stored on disk.

The second method of data augmentation is cropping. However, we have not used the strategy of random cropping for the fact of  low contrast between objects and background of UAV images, , since it may generate cropped image without object. Therefore, we manually crop five 600 x 600 images where the objects are in the four corners and the center, then,we make a precise label for each cropped image. 

With these two methods, we could prevent our network from overfitting. And before these images are sent to the detection network as input, we have to further resize these images to a fixed size to avoid the network overloading for a lack of memory.

\subsection{Region Proposal Networks}
A Region Proposal Network (RPN) takes as input an image and outputs a set of coordinates for each object proposal with an objectness score. The RPN shares computaion with a fast R-CNN by sharing a common set of convolutional layers. In our study, we have investigated the VGG16 \cite{30} and Res101 \cite{13} as backbone networks which have 13 and 98 shareable convolutional layers, respectively.

\begin{figure}[htbp]
\centering
\includegraphics[height=4cm,width=0.4\textwidth]{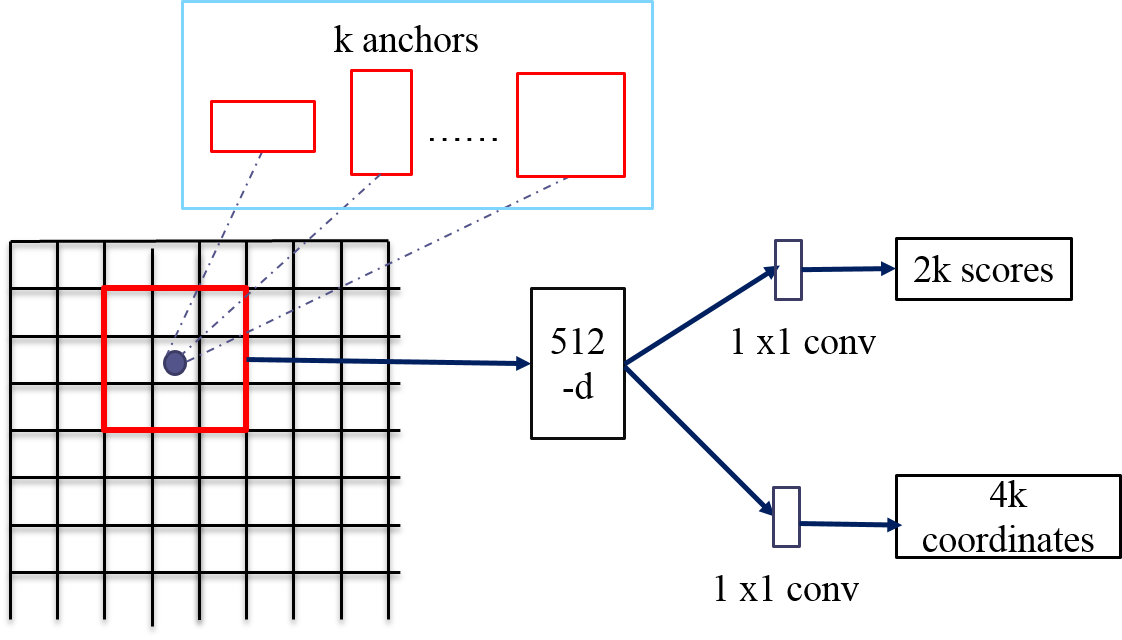}
\caption{\small mini-network of region proposals network}
\label{fig2}
\end{figure}

To generate region proposals, a small network is slid over the feature maps output by the last shared convolutional layer without max pooling. This small network takes as input an n x n (default n=3) spatial feature window and each feature window is turned into a 512 dimensional feature with ReLU \cite{31} following, then the feature is fed into two parallel fc layers for classification and bounding box regression. The detailed architecture of this mini-network is illustrated in Figure \ref{fig2}.

\paragraph{Anchors} While sliding the window over the feature map, RPN simultaneously predicts k region proposals also called anchors at each window location. So the cls layer outputs 2k scores that estimate probability of object or background for each anchors and the reg layer has 4k outputs encoding the coordinates of k anchors. An anchor is centered at the sliding window with a scale and aspect ratio.

\paragraph{Loss Function} Before training the RPN, we assign a binary class label to each anchor. An anchor that has an Intersection-over-Union (IoU) overlap higher than 0.7 with any ground-truth box is assigned to a positive label while a negative label is assigned to the anchor whose IoU ratio is lower than 0.3 for all ground-truth boxes. Anchors that are neither positive nor negative do not contribute to the training loss function.

\qquad With these definitions, RPN can be trained to minimize a multi-task loss function which is defined as:
\begin{equation}
\begin{split}
L(\left \{ p_{i}\right \},\left \{t_{i}\right \})=\frac{1}{N_{cls}} \sum_{i}^{} L_{cls}\left ( p_{i}p_{i}^{*} \right )\\+\lambda \frac{1}{N_{reg}}\sum_{i}^{j}p_{i}^{*}L_{reg}\left ( t_{i},t_{i}^{*} \right ) \label{eq}
\end{split}
\end{equation}
Here, i is the index of an anchor and \(p_{i}\) is the predicted probability of anchor i being an object. The ground-truth label \(p_{i}^{*}\) is 1 if the anchor is positive, otherwise 0. \(t_{i}\) is a vector representing the 4 parameterized coordinates of the predicted bounding box, and \(t_{i}^{*}\) is that of the ground-truth box associated with a positive anchor. The classification loss \(L_{cls}\) is log loss over two classes. For the regression loss, \(L_{reg}\left ( t_{i},t_{i}^{*} \right )\) is the robust loss function called smooth L1 defined in \cite{14}. The two terms are normalized by \(N_{cls}\) and \(N_{reg}\) and weighted by a balancing parameter \(\lambda\). \(N_{cls}\) indicates the mini-batch size and \(N_{reg}\) is the number of anchor locations. \(\lambda=10\) by default and thus both cls and reg term are roughly equally weighted.

\paragraph{Training of RPNs} Region Proposals Networks can be trained end-to-end by backpropagation and stochastic gradient descent (SGD) \cite{32}. RPN randomly sample 256 anchors in an image to compute the loss function of a mini-batch, where the sampled positive and negative anchors have a ratio of up to 1:1.

\quad All the new layers of RPN are randomly initialize by a zero-mean Gaussian distribution with standard deviation 0.01 and the shared convolutional layers are initialized by a pre-trained model in ImageNet \cite{33}.

\subsection{Fast R-CNN}
A fast R-CNN takes an entire image and a set of object proposals as input. Fast R-CNN first extracts feature map from the image by a set of convolutional and max pooling layers. Then, each region proposal is projected to the feature map. Instead of using the ROI Pooling layer in \cite{14}, we use crop and resize operation followed by a max pooling layer to generate a fixed size RoI feature map for the subsequent fully connected layers. The operation which uses bilinear interpolation to crop and resize a fixed size feature map is implemented in Tensorflow. Each resized RoI feature map is input into a set of fc layers that finally branch into two sibling output layers: one that produces softmax probability estimates over K object classes plus a “background” class and another layer outputs four real-valued numbers which encode refined bounding-box positions for one of the K classes. 

\paragraph {Loss Function}
Fast R-CNN also uses multi-task loss function to jointly train for classification and bounding-box regression:
\begin{equation}
\begin{split}
 L\left ( p,u,t^{u},v \right )=L_{cls}\left ( p,u \right )+\\
 \lambda \left [ u \geq 1 \right ]L_{reg}\left ( t^{u},v \right )\label{eq}
\end{split}
\end{equation}

Here, p is the softmax probability of a RoI over K+1 categories, and \(L_{cls}\left ( p,u \right )\) is log loss for class u. Regression loss \(L_{reg}\left ( t^{u},v \right )\)  is defined over a  tuple of true bounding-box regression targets for class u, the ground-truth bounding-box regression target \(v=(v_{x} , v_{y} ,v_{w} ,v_{h})\) and the predicted target \( t^{u}=\left ( t_{x}^{u}, t_{y}^{u}, t_{w}^{u}, t_{h}^{u} \right )\) which have the same defination as \eqref{reg}. \(\left [ u \geq 1 \right ]\) means that background RoI is ignored for bounding box regression. \(L_{reg}\left ( t^{u},v \right )\) is a robust L1 loss function:
\begin{equation}
\begin{split}
L_{reg}\left ( t^{u},v \right )=\sum_{i\in \left ( x,y,w,h \right )}smooth_{L_{1}}\left ( t_{i}^{u}-v_{i} \right )\label{eq}
\end{split}
\end{equation}
in which 
\begin{equation}
\begin{split}
smooth_{L_{1}}\left ( x \right )=\left\{\begin{matrix}
0.5x^{2} & if\left | x \right |< 1\\ 
\left | x \right |-0.5 & otherwise
\end{matrix}\right. \label{eq}
\end{split}
\end{equation}
The	hyper-parameter \(\lambda\) controls the balance between the two task losses and \(\lambda=1\) by default.

\paragraph{Training Fast R-CNN}The backbone network of fast R-CNN is initialized with a pre-trained ImageNet network which is the same as RPN, and the other new layers are initialized with a zero-mean Gaussian distribution with standard deviation 0.01. Then we fine-tune the whole detection network.

\subsection{Sharing features for RPN and fast R-CNN}
Like Faster R-CNN, we adopt the way of 4-Step Alternating Training to learn a unified detection network composed of RPN and fast R-CNN with shared convolutional layers, as shown below:
\begin{enumerate}
    \item Training RPN: RPN is initialized with a pre-trained model of ImageNet such as VGG16 and fine-tuned end-to-end for the region proposal task. 
    \item Training fast R-CNN: a separate fast R-CNN is also initialized with a pre-trained model which is the same as RPN, and takes as input the region proposals generated by the step-1 RPN.
    \item Initialize the training of RPN with the trained fast R-CNN, but fix the shared convolutional layers and only fine-tune the layers unique to RPN.
    \item Keeping the shared convolutional layers fixed, and fine-tune the unique layers of fast R-CNN.
\end{enumerate}
\qquad In this way, both networks have the same convolutional layers and become a unified detection network.

\subsection{Implementation Details}
In our proposed method, the detection network is trained and tested on single size images \cite{14}, \cite{27}.We first resize the images such that their shorter side reaches at a fixed size. Then, the larger side is resized at the same ratio with the maximum size threshold.This operation is important to avoid network overloading for the high resolution of UAV images. For anchors, we use 3 scales and 3 aspect ratios which both are the same with original faster R-CNN. We ignore all the anchors that cross image boundaries so they do not contribute to the loss during training. While during testing, we keep all cross-boundary anchors which we clip to the image boundary.

Region proposals network generates tens of thousand of anchors, however, these anchors highly overlap with each other. To reduce redundancy, we employ non-maximum suppression (NMS) on the anchors based on their cls scores and set the IoU threshold for NMS at 0.7. As reported in [20], the NMS does not harm the ultimate detection accuracy but increases the detection speed.

During training, we use a learning rate of 0.001 for the first 40k iteration and 0.0001 for the rest on the UAV image training dataset with a momentum of 0.9 and a weight decay of 0.0005\cite{12}. Our implementation is based on Tensorflow.

\begin{figure*}

  \centering 
  \subfloat{
    \includegraphics[width=0.5\textwidth]{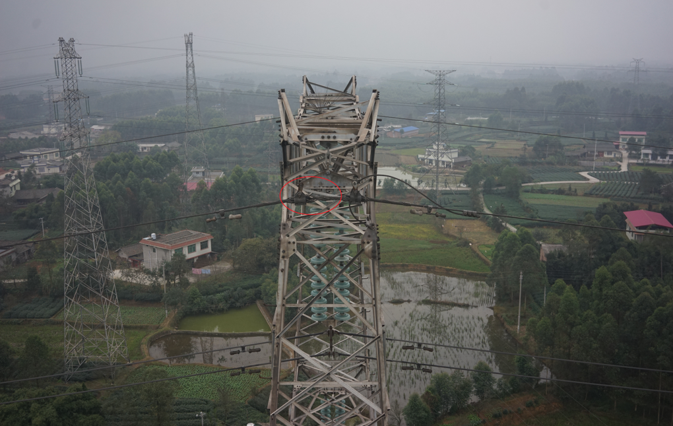} 
  } 
  \subfloat{ 
    \includegraphics[width=0.5\textwidth]{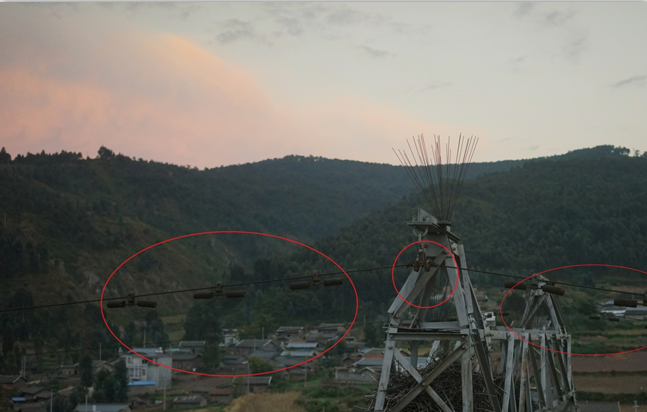} 
  } 
 \caption{The original UAV images of electrical equipment with various scales and complex background.} 
 \label{fig3}
\end{figure*}

\section{experiments}
\subsection{Dataset description}
In this study, we have collected a lot of multi-class detective electrical equipment images taken by UAV. After filtrating the images that are blurry and dark, we have labeled four classes of images including the pole and tower, wire, electric fittings and insulator,respectively. The training set consists of 2000 images of pole and tower, 1800 images of electric fittings, 1200 images of insulator and 600 images of wire. As we can observe that the numbers of the four classes are imbalanced, because actually some different defects of pole and \& tower are grouped into the same class. The testing set has 200 images in total and 50 images per class, as shown in Table \ref{tab1}. For a more intuitive understanding, Figure \ref{fig3} shows some original UAV images of electrical equipment with various scales and complex background. 

\begin{table}[htbp]
\caption{Details of the electrical equipment dataset}
\begin{center}
\begin{tabular}{ccc}
\hline
Type& Training set& Testing set\\
\hline
Insulator&1200&50\\
Pole and tower&2000&50\\
Fitting&1800&50\\
Wire&600&50\\
\hline
In total&5600&200\\
\hline
\end{tabular}
\label{tab1}
\end{center}
\end{table}
\subsection{Experimental Results}

\paragraph{Experiments respect to backbone networks} We evaluate the performance of the modified detection model with two different pre-trained backbone networks,  i.e., VGG16 and Resnet101, and Table \ref{tab2} shows the results of VGG16 and resnet101 at the same short side scale.Using VGG16 as the backbone network has the mAP of 81.11\% and Resnet has the mAP of 67.92\%. However, this is an unusual result compared with the network testing on conventional dataset such as PASCAL VOC [34] and COCO [35] on which resnet101 has a better performace than VGG16 in detection tasks. According to these preliminary experimental results, we analyze that the very deep CNN resnet101 may overfit to our training set for the imbalanced and insufficient samples in each class. According to the results of AP for each class, pole and tower has best performance because enough training samples and relatively obvious defective object. It cost about 0.3s per image at the scale of 1000 in inference.

\begin{table}
\caption{Results on the testing set with modified region-based CNN of two backbone networks}
\begin{center}
\begin{tabular}{ccccccc}
\hline
Backbone net& scale & mAP & insulator & tower & fitting & wire\\
\hline
VGG16 & 1000 & 81.11 & 87.87 & 93.16 & 75.12 & 68.28\\
ResNet101 & 1000 & 67.92 & 73.81 & 80.48 & 72.91 & 44.48\\
\hline
\end{tabular}
\label{tab2}
\end{center}
\end{table}

\paragraph{Experiments respect to scale of input image}
For the reason that we have to resize the high resolution images to a suitable scale to prevent our detection model form overloading. Besides, considerable defective objects in the UAV images are very small and unobvious because of the distance and the angle of the UAV images, so the resized scale of input image decides the receptive filed of the detection model for these small objects. We evaluate there scales for the input image on detection model with two different backbone networks, respectively. Table \ref{tab3} shows the results. Form the results, we can see that our detection model gets a better performance with the increase of input image size. The input scale has great influence on the small defective object detection, but it also increases the inference time.

\begin{table}
\caption{Results on the testing set with modified region-based R-CNN of two backbone networks}
\begin{center}
\begin{tabular}{ccccccc}
\hline
Backbone net& scale & mAP & insulator & tower & fitting & wire\\
\hline
VGG16 & 1000 & 81.11 & 87.87 & 93.16 & 75.12 & 68.28\\
VGG16 & 800 & 65.88 & 67.84 & 86.22 & 71.92 & 35.55\\
VGG16 & 600 & 63.00 & 65.84 & 84.45 & 66.26 & 34.45\\
\hline
ResNet101 & 1000 & 67.92 & 73.81 & 80.48 & 72.91 & 44.48\\
ResNet101 & 800 & 55.37 & 66.83 & 71.59 & 61.09 & 21.97\\
ResNet101 & 600 & 46.27 & 50.15 & 69.80 & 56.05 & 9.09\\
\end{tabular}
\label{tab3}
\end{center}
\end{table}

\begin{figure*}
\centering 
  \subfloat{
    \includegraphics[scale=0.2]{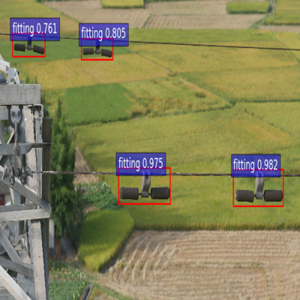} 
  } 
  \subfloat{ 
    \includegraphics[scale=0.2]{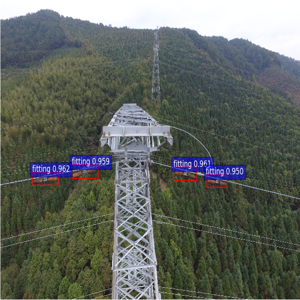} 
  } 
 \subfloat{
    \includegraphics[scale=0.2]{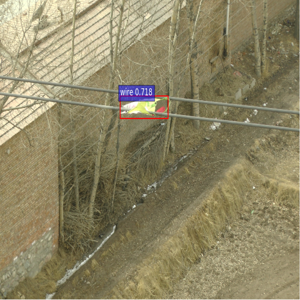} 
  } 
  \subfloat{ 
    \includegraphics[scale=0.2]{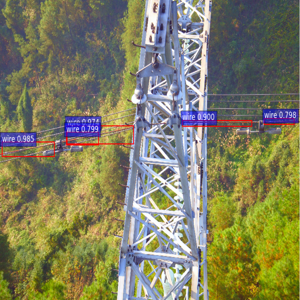} 
  }
  
   \subfloat{
    \includegraphics[scale=0.2]{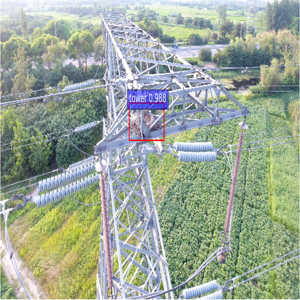} 
  } 
  \subfloat{ 
    \includegraphics[scale=0.2]{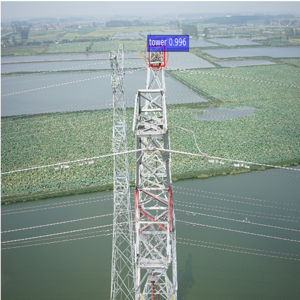} 
  } 
 \subfloat{
    \includegraphics[scale=0.2]{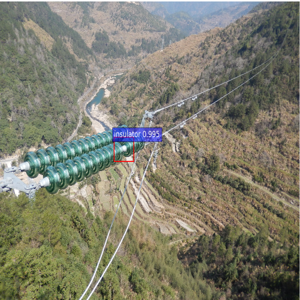} 
  } 
  \subfloat{ 
    \includegraphics[scale=0.2]{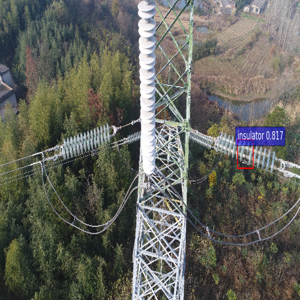} 
  } 
\caption{ Selected resized examples of object detection results on our testing set using the region-based CNN detection model.The backbone network is VGG16 and scale is 1000. Each object box is associated with a category label and a softmax score}  
 \label{fig4}
\end{figure*}

\paragraph{Visualization of Defect Detection}
In this section, we visualize some our inference samples for the four class and visually show the feasibility of applying detection methods based on deep learning to detect multiple class defective electrical equipments. As shown in Figure \ref{fig4}

\section*{CONCLUSION AND FUTURE WORK}
In this work, we apply a modified region-based CNN to a new application of defect detection of electrical equipment. Based on the properties of UAV HD images, we preprocess the image data to avoid  overfitting. The major modifications of the detection network lie in that (1) we have replaced the RoI pooling layer with crop\_and\_resize operation which is fast and easily implemented on Tensorflow, and (2) we have used R=256, N=1 instead of R=128 and N=2, in order to address the issue of slow training convergence. We have also considered the feature pyramid for small object detection. Our experimental results indicate that the region-based CNN can well detect the defects of multiple classes of electrical equipment and is significantly more efficient and accurate than the traditional single-class detection methods.

\end{document}